\documentclass[10pt,twocolumn]{article}

\usepackage{graphicx}
\usepackage{algorithm,algorithmic}
\usepackage{amsmath}
\usepackage{multirow}
\usepackage{float}
\usepackage{url}
\usepackage[none]{hyphenat}
\usepackage{authblk}

\usepackage{float}
\newcommand\T{\rule{0pt}{2.6ex}}
\newcommand\B{\rule[-1.2ex]{0pt}{0pt}}
\newcommand\TT{\rule{0pt}{3.0ex}}
\newcommand\BB{\rule[-1.5ex]{0pt}{0pt}}

\begin{document}
\newcounter{cntr1}
\newcounter{cntr2}
\emergencystretch 3em

\title{Water Distribution System Design Using Multi-Objective\\
Particle Swarm Optimisation}

\author[1]{Mahesh B. Patil}
\author[2]{M. Naveen Naidu}
\author[2]{A. Vasan}
\author[2]{Murari R. R. Varma}
\affil[1]{Department of Electrical Engineering, Indian Institute of Technology Bombay}
\affil[2]{Department of Civil Engineering, BITS Pilani Hyderabad Campus}

\maketitle

\begin{abstract}
Application of the multi-objective particle swarm optimisation
(MOPSO) algorithm to design of water distribution systems is
described. An earlier MOPSO algorithm is augmented with
(a)~local search, (b)~a modified strategy for assigning
the leader, and (c)~a modified mutation scheme. For one of
the benchmark problems described in the literature, the
effect of each of the above features on the algorithm performance
is demonstrated. The augmented MOPSO algorithm (called MOPSO+)
is applied to five benchmark problems, and in each case, it finds
non-dominated solutions not reported earlier.
In addition, for the purpose of comparing Pareto fronts (sets
of non-dominated solutions) obtained by different algorithms,
a new criterion is suggested, and its usefulness is pointed
out with an example. Finally, some suggestions regarding future
research directions are made.
\end{abstract}

\section{Introduction}
Design of water distribution systems (WDS) involves various
types of challenging optimisation problems which have been
investigated in several research papers to date\,%
\cite{farmani2005}-%
\nocite{janga2007}%
\nocite{kadu2008}%
\nocite{montalvo2010}%
\nocite{raad2011}%
\nocite{ahmadi2013}%
\nocite{wang2014}%
\nocite{barlow2014}%
\nocite{morley2014}%
\nocite{mora2015}%
\nocite{bi2016}%
\nocite{moosavian2016}%
\cite{zheng2017}.
An important subset of these problems is the two-objective
optimisation problem in which pipe diameters are to be determined
with the objectives of minimising the total cost and maximising
the network resilience. Recently, Wang {\it{et al.}}\,\cite{wang2014}
have reported a systematic collection of twelve benchmark WDS
design problems of various complexities to serve as an excellent
starting point for researchers to try out new multi-objective
optimisation algorithms and compare their performance with the
results available in the literature. Furthermore, they have made
the corresponding EPANET\,\cite{rossman2000} source code and the best-known
Pareto front data available in the public domain\,\cite{exeter}.

\begin{table}
   \centering
   \caption{Summary of some of the benchmark WDS design problems
    presented in \cite{wang2014}.}
    \hspace*{0cm}
    \begin{tabular}{|c|c|c|c|c|c|}
    \hline
    {\multirow{2}{*}{Network}} &
    {\multirow{2}{*}{$N_ {\mathrm{pipes}}$}} &
    {\multirow{2}{*}{$N_ {\mathrm{dia}}$}} &
    Search &
    {\multirow{2}{*}{$N_{FE}^ {\mathrm{net}}$}} &
    {\multirow{2}{*}{$N_{PF}$}} \\
    {} & {} & {} & space size & {} & {} \\ \cline{1-6}
    \hline
    TLN\B\T\,(SP) &  8 & 14 & $1.48\times 10^9$     &  15\,M & 128 \\ \hline
    NYT\B\T\,(MP) & 21 & 16 & $1.93\times 10^{25}$  &  90\,M & 627 \\ \hline
    BLA\B\T\,(MP) & 23 & 14 & $2.30\times 10^{26}$  &  90\,M & 901 \\ \hline
    HAN\B\T\,(MP) & 34 &  6 & $2.87\times 10^{26}$  &  90\,M & 575 \\ \hline
    GOY\B\T\,(MP) & 30 &  8 & $1.24\times 10^{27}$  &  90\,M & 480 \\ \hline
    PES\B\T\,(IP) & 99 & 13 & $1.91\times 10^{110}$ & 150\,M & 782 \\ \hline
    \end{tabular}
\label{tbl_wang1}
\end{table}

Of the twelve benchmark problems presented in \cite{wang2014}, six are
summarised in Table~\ref{tbl_wang1}.
For a ``small problem" (SP), the size of the search space is relatively
small, e.g., for the TLN problem, it is
$14^8 \,$=$\, 1.48\times 10^9$,
and it is possible to compute the cost and resilience for each of the
$14^8$ possible networks to obtain the true Pareto front. In larger
problems, including medium and intermediate problems (marked as MP
and IP, respectively, in the table), it is not possible to compute the
objective function values for each possible network configuration, and
a multi-objective evolutionary algorithm (MOEA) is employed to obtain
the approximate Pareto front (i.e., set of non-dominated solutions).
In these cases, the true Pareto front is not known, but
following \cite{wang2014}, we will often use the term ``Pareto front" (PF)
to refer to the set of non-dominated (ND) solutions obtained by the
concerned algorithm.

Five MOEAs were used in \cite{wang2014} for each of the benchmark problems.
The ND solutions obtained by the five algorithms were combined, and
dominated solutions were removed to form the ``best-known" PF.
The number of solutions ($N_{PF}$) in the best-known PF thus obtained
is listed in Table~\ref{tbl_wang1} for each benchmark problem.
The percentage of solutions contributed to $N_{PF}$ for each problem
by each of the five algorithms is given in Table~\ref{tbl_wang2}.
It can be observed that, except for TLN and HAN networks, no single
algorithm was able to obtain all of the solutions in the best-known PF.

\begin{table*}
   \centering
   \caption{Percentage of solutions contributed by five MOEAs to the
    best-known Pareto front\,\cite{wang2014}.}
    \hspace*{0cm}
    \begin{tabular}{|c|r|r|r|r|r|}
    \hline
    {\multirow{2}{*}{Network}}
    & \multicolumn{5}{c|}{\B \T \% contribution}\\
    \cline{2-6}
    {} &
          \multicolumn{1}{c|}{\B\T NSGA-II} &
          \multicolumn{1}{c|}{$\epsilon$-MOEA} &
          \multicolumn{1}{c|}{$\epsilon$-NSGA-II} &
          \multicolumn{1}{c|}{AMALGAM} &
          \multicolumn{1}{c|}{Borg}
    \\ \hline
    TLN\B\T\,(SP) & 100.0~~ &  83.1~~~ &  83.1~~~~ &  98.7~~~~~~ & 84.4~ \\ \hline
    NYT\B\T\,(MP) &  93.8~~ &  17.9~~~ &  24.8~~~~ &  91.7~~~~~~ & 20.7~ \\ \hline
    BLA\B\T\,(MP) &  77.0~~ &  30.3~~~ &  26.3~~~~ &  90.1~~~~~~ & 28.3~ \\ \hline
    HAN\B\T\,(MP) & 100.0~~ &  20.5~~~ &  23.1~~~~ &  89.7~~~~~~ & 25.6~ \\ \hline
    GOY\B\T\,(MP) &  43.3~~ &   3.0~~~ &  58.2~~~~ &  85.1~~~~~~ & 22.4~ \\ \hline
    PES\B\T\,(IP) &  38.1~~ &  27.0~~~ &  10.7~~~~ &  18.6~~~~~~ & 23.3~ \\ \hline
    \end{tabular}
\label{tbl_wang2}
\end{table*}

The computation effort involved in obtaining the best-known
PF can be described in terms of the total number of function
evaluations
$N_{FE}^{\mathrm{net}}$
by all five algorithms together since no single algorithm was
found adequate in general. For the problems in the MP category,
$N_{FE}$ was restricted to 600,000 for each of the five algorithms\,\cite{wang2014},
and 30 independent (randomly initiated) runs were performed for each
benchmark problem. The total
$N_{FE}$ is therefore
$N_{FE}^{\mathrm{net}} \,$=$\, 600,000\times 30 \times 5 \,$=$\, 90$\,million (90\,M),
as shown in Table~\ref{tbl_wang1}.

The purpose of this paper is to present a modified
multi-objective particle swarm optimisation algorithm
(MOPSO+) for WDS design. In particular, we present
results for some of the benchmark problems described in
\cite{wang2014} and show that the proposed algorithm gives
better solutions than the best-known PFs of \cite{wang2014}.
The paper is organised as follows. In Sec.~\ref{sec_compare_pf},
we present a new scheme for comparing PFs produced by two
algorithms, to be used in subsequent sections to gauge the
performance of the MOPSO+ algorithm. In Sec.~\ref{sec_ls},
we describe the local search scheme implemented in MOPSO+.
In Sec.~\ref{sec_mopso}, we describe the overall MOPSO+ algorithm
and point out specifically the changes with respect to the original MOPSO
algorithm presented in \cite{coello2004}. We compare the results
obtained using MOPSO+ with the best-known PFs for medium problems
in Sec.~\ref{sec_medium}. We then point out in Sec.~\ref{sec_inter}
limitations of the local search scheme for larger problems,
propose a suitable modification, and show its effectiveness
for one of the benchmark problems in the IP category\,\cite{wang2014}.
Finally, we present conclusion of this work in Sec.~\ref{sec_concl}
and comment on related future work.

\section{Comparison of Pareto fronts}
\label{sec_compare_pf}
In comparing the performance of an MOEA with another with respect
to a given benchmark problem, we are mainly interested in
(a)\,computation time, (b)\,the ``quality" of the set of
ND solutions. Computation time depends on several factors such as
the hardware used, programming environment (compiled or interpreted),
and whether parallelisation was employed. Because these factors would
vary, a more objective metric, viz., the total number of function
evaluations
$N_{FE}^{\mathrm{net}}$
is used in \cite{wang2014} and also in this work.

To judge the quality of the ND set obtained by an algorithm, the
following metrics are commonly used\,\cite{coello2004},\cite{kdeb1}
when the decision variables are
continuous: (a)\,generational distance, a measure of how far the
obtained ND solutions are from the true Pareto front, (b)\,spacing,
a measure of the spread of solutions throughout the ND set,
(c)\,error ratio, the percentage of the ND solutions which do not
belong to the true Pareto-optimal set.

In the context of the benchmark WDS problems considered here, the
true Pareto front is not known. In any case, the above metrics are
of limited value for the WDS problem because of the discrete nature
of the decision variables. From a practical perspective, apart from
cost and resilience, the decision maker (DM) may need to take into
account other considerations for implementation\,\cite{montalvo2010},
and it is important to make a large number of options available to
the DM. A more meaningful metric is therefore the number of solutions
in the best-known PF which the concerned algorithm can
contribute\,\cite{wang2014}.
Based on this idea, we propose some new metrics for comparing the
PFs obtained by two MOEAs for a given benchmark problem.

\begin{figure*}[!ht]
\centering
\scalebox{0.9}{\includegraphics{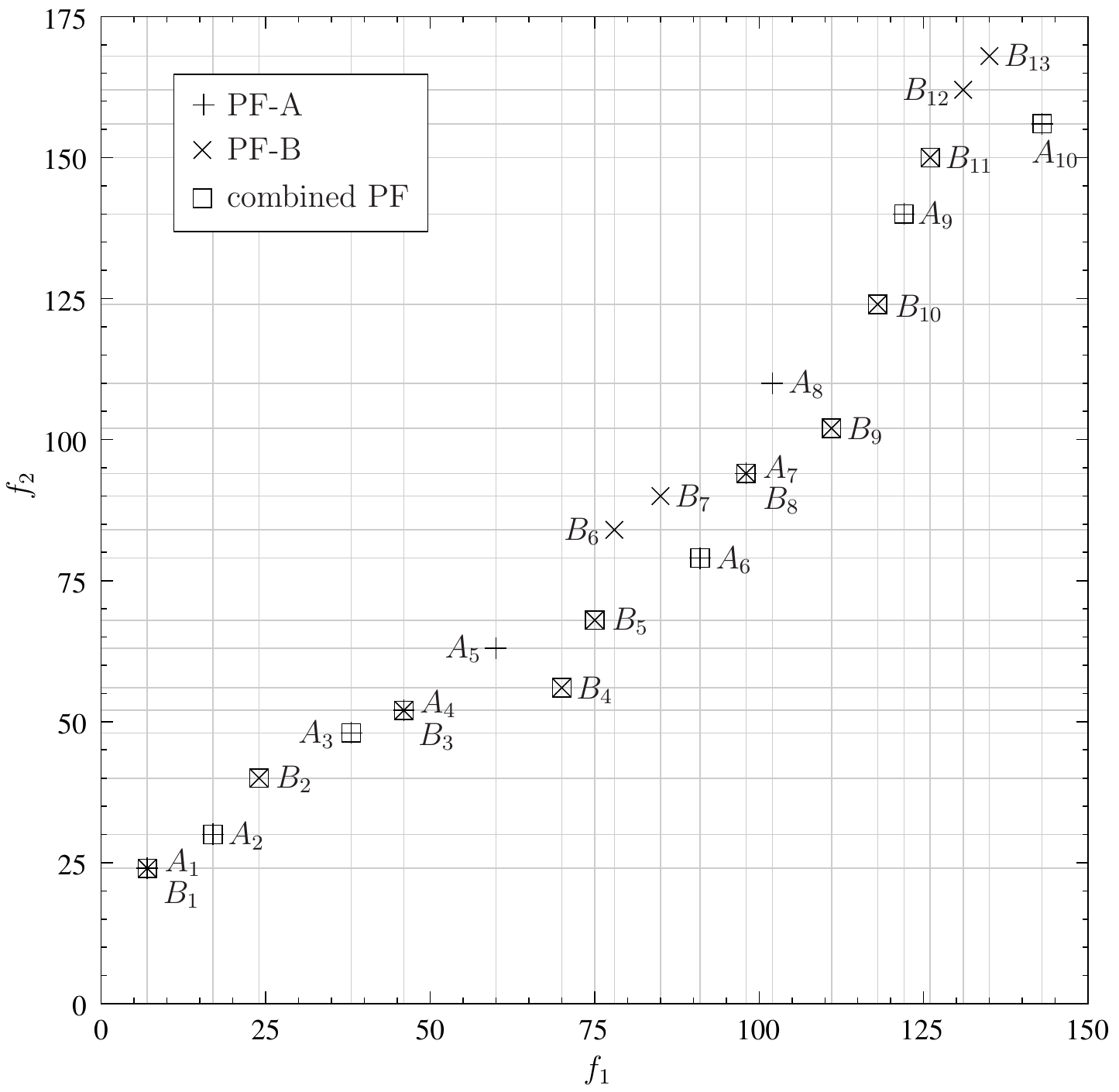}}
\caption{Example showing two sets of ND solutions (Pareto
fronts) A and B.}
\label{fig_metric}
\end{figure*}

Consider a two-objective problem in which the objective functions
$f_1$ and $f_2$ are to be maximised and minimised, respectively.
Fig.~\ref{fig_metric} shows an example with two sets of ND solutions
referred to as PF-A and PF-B. The combined set of ND solutions
(marked as ``combined PF" in the figure) is obtained by combining
solutions from PF-A and PF-B and then removing solutions dominated
by any other solutions. The combined PF contains 14 solutions. PF-A,
which contains a total of 10 solutions ($A_1$ to $A_{10}$), contributes
5 unique solutions
($A_2$, $A_3$, $A_6$, $A_9$, $A_{10}$), i.e., solutions which are not
present in PF-B.
Similarly, PF-B,
which contains a total of 13 solutions ($B_1$ to $B_{13}$), contributes
6 unique solutions
($B_2$, $B_4$, $B_5$, $B_9$, $B_{10}$ $B_{11}$) to the combined PF.
The number of common solutions in the combined PF, i.e., solutions
which are contributed by PF-A as well as PF-B, is 3
($A_1$/$B_1$, $A_4$/$B_3$, $A_7$/$B_8$). Note that some of the solutions
have been rejected from the original PFs, viz., 2 solutions
($A_5$, $A_8$) from PF-A and 4 solutions
($B_6$, $B_7$, $B_{12}$ $B_{13}$) from PF-B.

We use the following nomenclature to describe the above scenario:
$N_A^t \,$=$\, 10$,
$N_A^u \,$=$\, 5$,
$N_A^c \,$=$\, 3$,
$N_A^r \,$=$\, 2$
for PF-A, and
$N_B^t \,$=$\, 13$,
$N_B^u \,$=$\, 6$,
$N_B^c \,$=$\, 3$,
$N_B^r \,$=$\, 4$
for PF-B.
The superscripts t, u, c, r stand for total, unique, common, and
rejected, respectively. Note that
$N_A^c$ and
$N_B^c$ must be equal by definition, and we will denote them by
$N^c$. Clearly, for this example, we cannot say that PF-A is better
than PF-B (or {\it{vice versa}}) since each of them has contributed
some unique solutions to the combined set.

Consider now the case
$N_A^u \,$=$\, 0$,
$N_B^u \,$=$\, 4$.
In this case, PF-B is certainly better than PF-A because it contributes
some unique solutions while PF-A contributes none. In other words, we
can do without the information provided by PF-A since PF-B provides all
of the solutions which PF-A would have contributed to the combined front
($N^c$ of them), and $N_B^u$ additional solutions. We will use this concept
to compare the PFs obtained by the proposed MOPSO+ algorithm with the
best-known PFs\,\cite{exeter}.

\section{Local search implementation}
\label{sec_ls}
Memetic algorithms (MA), which combine the exploration capability
of an evolutionary algorithm with the exploitation capability of a
local search method, have been extensively used for multi-objective
optimisation\,%
\cite{ishibuchi2002}-%
\nocite{knowles2005}%
\nocite{sindhya2009}%
\nocite{mousa2012}%
\nocite{liefooghe2012}%
\nocite{dubois2013}%
\nocite{inja2014}%
\nocite{maler2016}%
\nocite{mansour2017}%
\cite{jaszkiewicz2018}.
Barlow and Tanyimboh\,\cite{barlow2014}
have presented a memetic algorithm for WDS optimisation. In their
work, the genetic algorithm (GA) was used along with local improvement.
The child population after every $N_G$ GA iterations was obtained
with local and cultural improvement operators using the following
procedure.
\begin{list}{(\alph{cntr2})}{\usecounter{cntr2}}
 \item
  Choose a subset S of the current ND set.
 \item
  Select one individual from S for local improvement.
  Define a scalar fitness function with linear weighting where the
  weights are obtained using an estimate of the gradient of the
  PF. For the selected individual,
  \begin{list}{(\roman{cntr2})}{\usecounter{cntr2}}
   \item
    Find the Hooke-Jeeves pattern search direction using the above
    scalar fitness function.
   \item
    Perform the cultural learning step by applying the same pattern
    search direction to a group of individuals in the current ND set.
  \end{list}
 \item
  Repeat (b) until a sufficient number of children are created.
\end{list}
The authors demonstrated that the introduction of local improvement
led to improved convergence speed and solutions better than previously
reported.

In this work, we use a simpler local search (LS) scheme based on the
observations that (a)\,New ND solutions can be found in the
neighbourhood of known ND solutions\,%
\cite{liefooghe2012},\cite{jaszkiewicz2018},
(b)\,Local search is effective in exploring the least-crowded areas of the objective
space (where a smaller number of ND solutions have been found so
far)\,\cite{mousa2012}. In the following, we illustrate the LS scheme used
in this work.

Consider the two-variable, two-objective test problem\,\cite{kita1996} in which
\begin{equation}
\begin{array}{lll}
f_1({\bf{x}})&=& -x_1^2+x_2\\
f_2({\bf{x}})&=&\frac{1}{2}\,x_1 + x_2 + 1
\end{array}
\label{eq_kita}
\end{equation}
are to be maximised, subject to the constraints specified in
\cite{kita1996}.

\begin{figure*}[!ht]
\centering
\scalebox{0.9}{\includegraphics{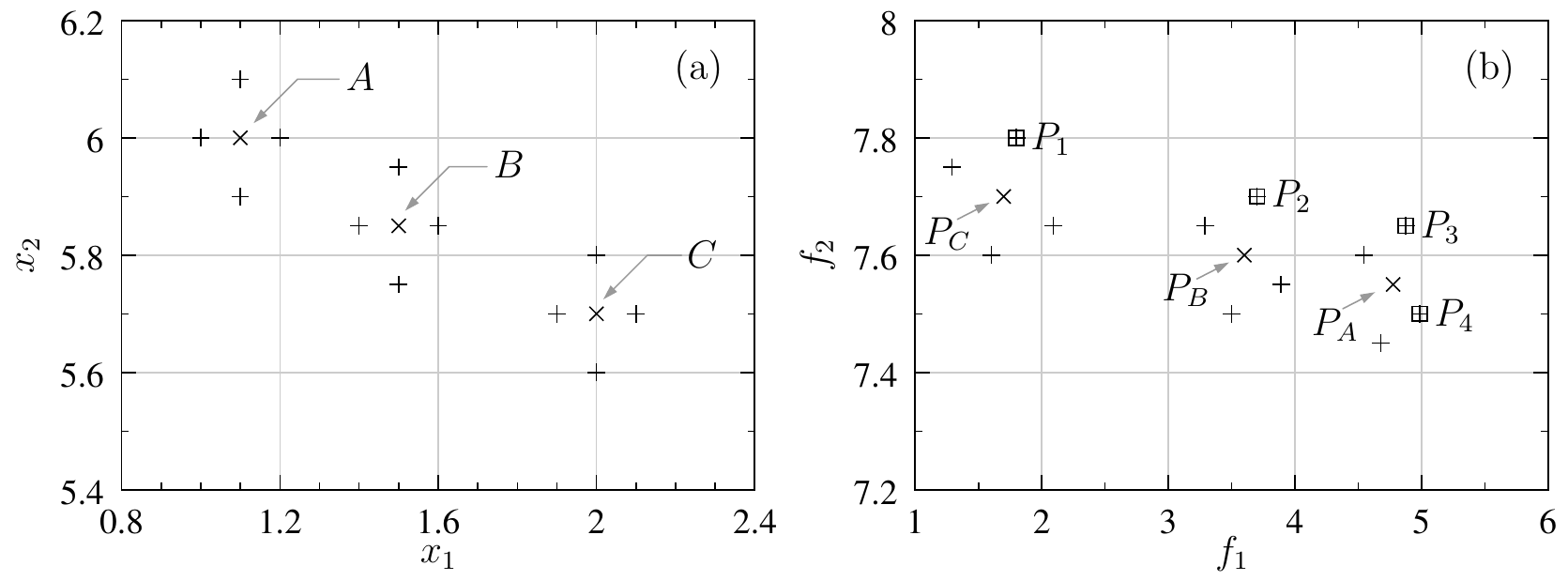}}
\caption{Illustration of local search for the optimisation problem
given by Eq.~\ref{eq_kita}. (a)\, Points in decision space,
(b)\,corresponding points in objective space. The original ND points,
the points generated by local search, and the new ND points are marked
with crosses, pluses, and squares, respectively.}
\label{fig_ls1}
\end{figure*}

Suppose the current ND set contains three points $A$, $B$, $C$ in
Fig.~\ref{fig_ls1}\,(a).
The corresponding objective function values are shown by
$P_A$, $P_B$, $P_C$ in
Fig.~\ref{fig_ls1}\,(b).
Our goal is to improve the ND set using local search. To this end,
we generate four neighbouring points centred around each existing
solution, as shown in
Fig.~\ref{fig_ls1}\,(a).
The $x_1$ and $x_2$ positions of the neighbours are given by
\begin{equation}
x_1^{\mathrm{new}} = x_1^{\mathrm{old}} \pm \Delta x_1,~~
x_2^{\mathrm{new}} = x_2^{\mathrm{old}} \pm \Delta x_2.
\label{eq_ls1}
\end{equation}

The corresponding neighbours in the objective space are shown in
Fig.~\ref{fig_ls1}\,(b).
We now find the ND solutions from the entire set consisting of the
old solutions and the newly generated solutions. The new ND set is
made up of
$P_1$, $P_2$, $P_3$, $P_4$ in
Fig.~\ref{fig_ls1}\,(b).
Note that the ND set has improved in two ways: (a)\,The number of
solutions has increased from 3 to 4, (b)\,The ND points in the new
set dominate the old ND points~-- in this example, all of the old
ND points
($P_A$, $P_B$, $P_C$).

\begin{figure*}[!ht]
\centering
\scalebox{0.9}{\includegraphics{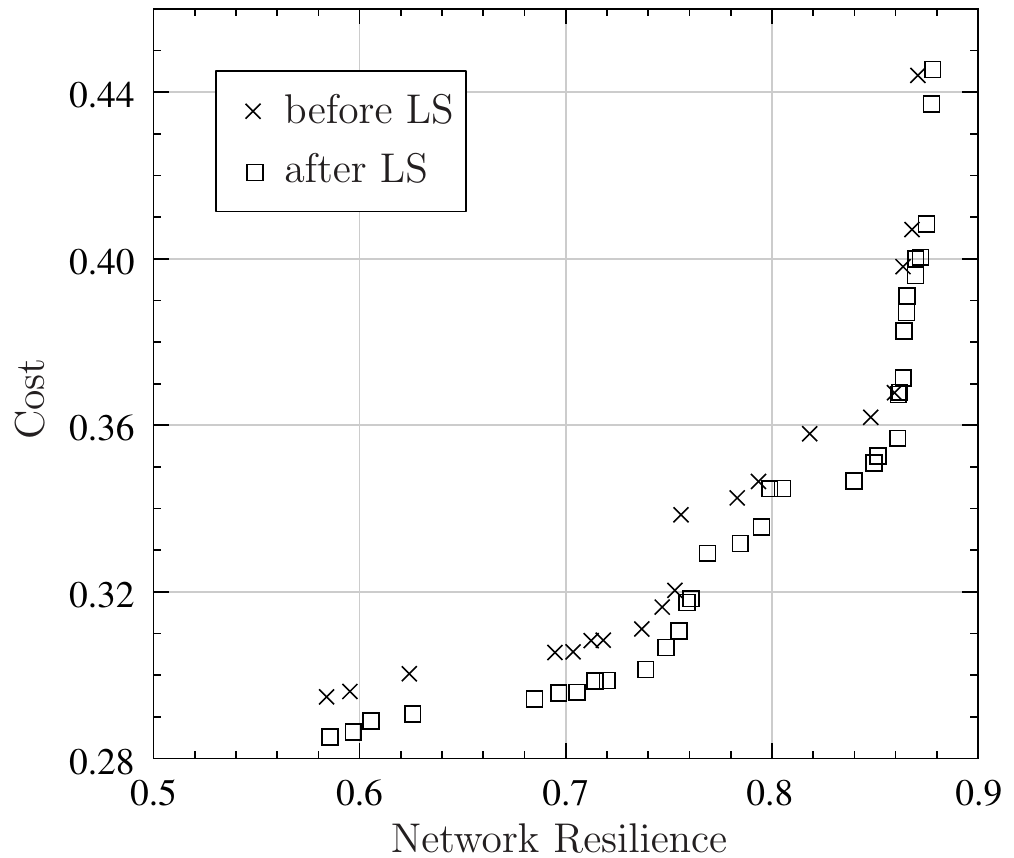}}
\caption{ND set for the BLA network at an early stage of the
MOEA before and after applying local search (cost is in millions
of USD).}
\label{fig_ls2}
\end{figure*}

It is important to evaluate the neighbours of each solution in the
ND set first and then update the ND set by removing dominated
solutions\,\cite{maler2016}.
For example, in Fig.~\ref{fig_ls1}\,(a), suppose we first evaluate
the neighbours of $A$ and obtain two new ND points
$P_3$ and $P_4$. If the ND set is updated at this stage, solution
$B$ will get removed (since $P_B$ is dominated by $P_3$).
As a consequence, the neighbours of $B$ will not get evaluated, and
the ND solution $P_2$ will not be obtained. To avoid this situation
in MOPSO+, we first prepare a list of all solutions whose neighbours
are to be explored. Subsequently, we treat the members of this list
one by one until the entire list is covered. The new solutions are
then used to update the ND set.

For the WDS problems we are considering here, the decision variables
are the pipe diameters. Each decision variable can take on values
1, 2, $\cdots$, $N_{\mathrm{dia}}$ (where
$N_{\mathrm{dia}}$ is the number of possible diameters, see Table~\ref{tbl_wang1}),
with 1 corresponding to the smallest diameter, 2 to the next larger diameter,
etc. For the BLA network, for example, there are 23 decision variables, each
taking a value between 1 and 14. We consider two solutions to be neighbours
if they differ by $\pm 1$ in one of the decision variables. In the BLA case,
therefore, a general solution not lying on a decision space boundary would
have $23\times 2 \,$=$\, 46$ neighbours. If there are $N'$ solutions in the
current ND set, we need to perform $46\times N'$ function evaluations, each
involving a call the the EPANET program for computation of cost and resilience.
The results before and after local search are shown in Fig.~\ref{fig_ls2}
for the BLA network at an early stage of the MOEA.
The improvement arising from local search can be clearly observed.

\begin{figure*}[!ht]
\centering
\scalebox{0.75}{\includegraphics{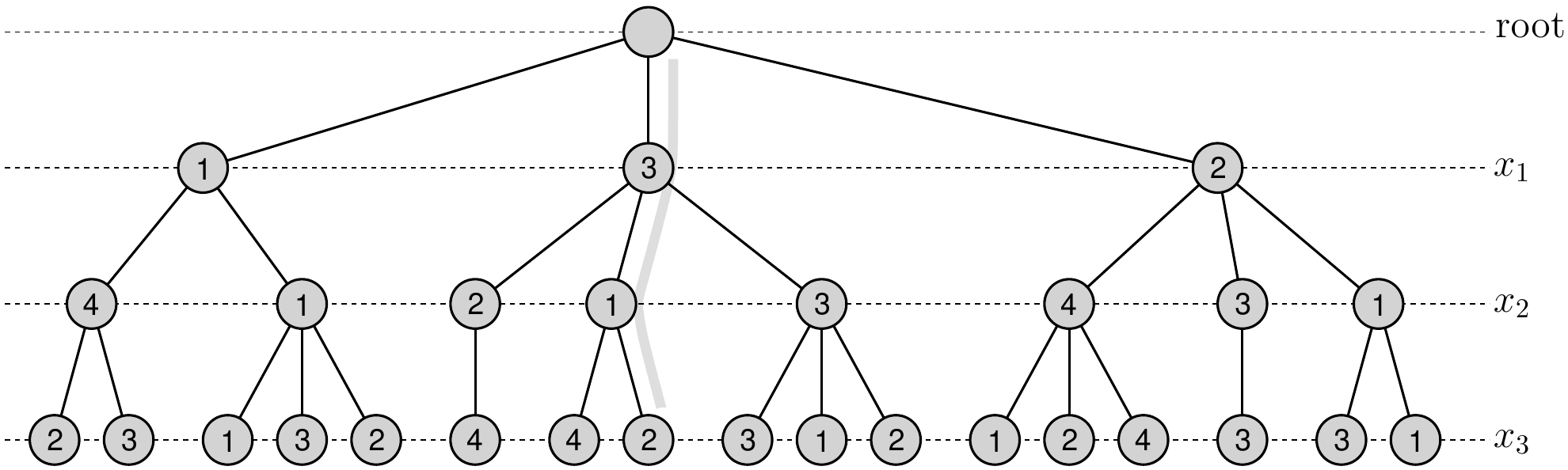}}
\caption{Tree structure used for storing solutions in the non-dominated set.}
\label{fig_tree}
\end{figure*}

The above procedure is obviously expensive. One way to reduce the
number of function evaluations is to keep track of points which have
already been evaluated. For these points, it is not necessary to
repeat function evaluation. However, comparison of a given point with a list
of points (in the decision space) is also expensive if it
is compared with each point in the list one by one. To make the comparison
more efficient, we employ a tree structure to represent the ND set.
An example is shown in Fig.~\ref{fig_tree}. In this case, there are
three decision variables $x_1$, $x_2$, $x_3$. Nodes at level $x_k$
are labeled by the $x_k$ values of the solutions. The path traced in
checking if the solution
$x_1 \,$=$\, 3$,
$x_2 \,$=$\, 1$,
$x_3 \,$=$\, 2$,
is present in the ND set is shown with a thick grey line.

\section{MOPSO+ algorithm}
\label{sec_mopso}
In this paper, we demonstrate the usefulness of a hybrid algorithm
using the MOPSO algorithm along with local search for the WDS design
problem. Several variations of the MOPSO algorithm have been reported
(see \cite{sierra2005},\cite{parsopoulos2008} for a review).
Here, we use the MOPSO algorithm
proposed by Coello\,\cite{coello2004}, with some modifications.

The velocity update formula in MOPSO is similar to that used in
single-objective PSO, and is given by
\begin{equation}
{\bf{v}}_i (t) =
W\,{\bf{v}}_i(t-1)
+ C_1 r_1({\bf{x}}_{\mathrm{pbest}}^i-{\bf{x}}_i (t))
+ C_2 r_2({\bf{x}}_{\mathrm{leader}}-{\bf{x}}_i (t)),
\label{eq_pso}
\end{equation}
where $t$ denotes the PSO iteration number,
${\bf{x}}_i$ and
${\bf{v}}_i$ are the position and velocity of the
$i^{\mathrm{th}}$ particle, respectively, and $r_1$ and $r_2$
are random numbers between 0 and 1. The algorithm parameters are
$W$ (inertia weight),
$C_1$ (cognitive learning factor), and
$C_2$ (social learning factor). In this work, we have used
$W \,$=$\, 0.4$,
$C_1 \,$=$\, 2$,
$C_2 \,$=$\, 2$. It should be pointed out that,
although the decision variables in the WSD benchmark problems
considered here take on discrete values, they are treated as real
numbers in the velocity and position update equations.
In computing the particle fitness, each decision variable is converted
to the nearest integer (which gives the pipe diameter index for the
concerned pipe).

\begin{figure*}[!ht]
\centering
\scalebox{0.6}{\includegraphics{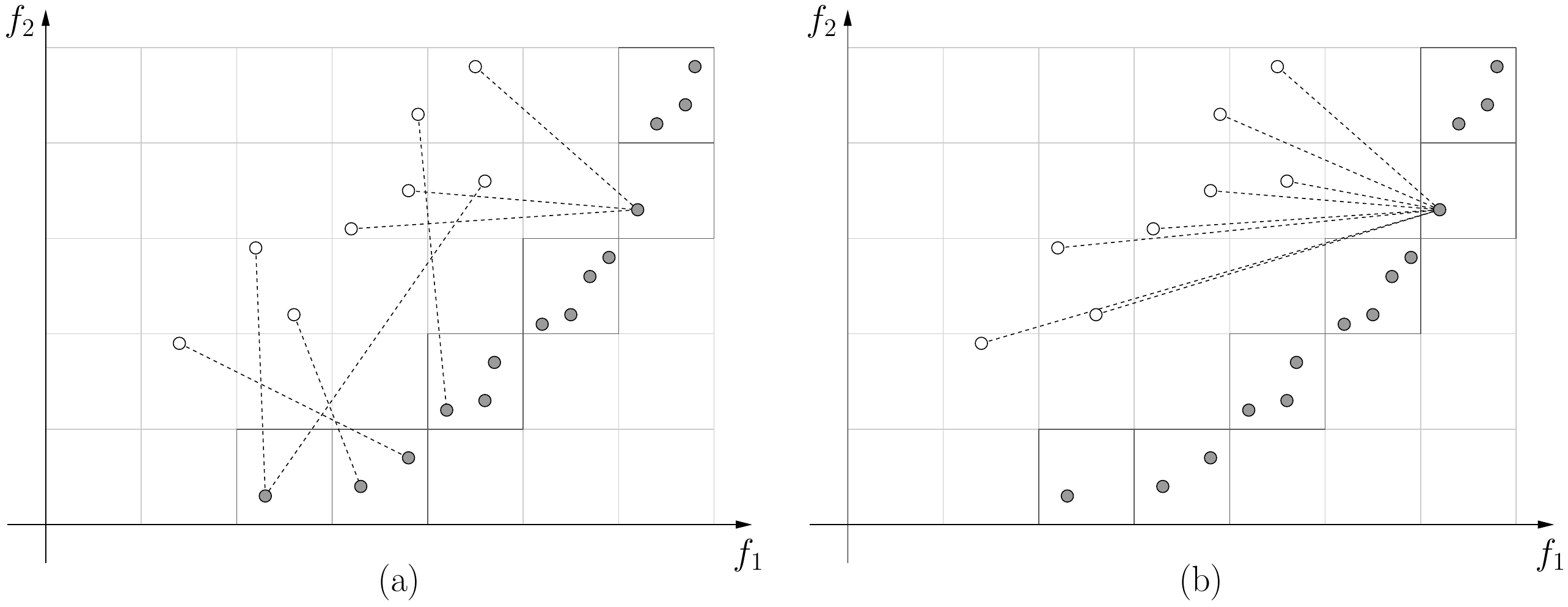}}
\caption{Illustration of leader selection procedure:
 (a)\,MOPSO algorithm\,\cite{coello2004}, (b)\,MOPSO+ algorithm (this work).
 Hollow circles: PSO particles, filled circles: current ND solutions.}
\label{fig_gbest}
\end{figure*}

The MOPSO algorithm differs from the single-objective PSO algorithm
in the computation of
${\bf{x}}_{\mathrm{pbest}}^i$
(the personal best position of the particle so far) and
${\bf{x}}_{\mathrm{leader}}$
(the position of the leader).
In \cite{coello2004},
${\bf{x}}_{\mathrm{pbest}}^i$
is updated in every iteration by comparing its current position
with the previous value of
${\bf{x}}_{\mathrm{pbest}}^i$.
If the current position dominates, it replaces
${\bf{x}}_{\mathrm{pbest}}^i$.
If it is non-dominated with respect to
${\bf{x}}_{\mathrm{pbest}}^i$, then one of them is selected randomly as
the next
${\bf{x}}_{\mathrm{pbest}}^i$.

In MOPSO\,\cite{coello2004}, assignment of 
${\bf{x}}_{\mathrm{leader}}$
is made using the ND set stored in an external archive
(repository). The objective space is divided into hypercubes, and
each ND solution, depending on its position in the objective space,
is assigned one of these hypercubes.
For each particle, in each
PSO iteration, a leader is selected from the archive giving preference
to ND solutions which occupy less-crowded hypercubes. Roulette-wheel
selection is used to first select a hypercube, and one of the ND solutions
in that hypercube is picked randomly as the leader. This procedure helps
to ensure that the ND solutions are well distributed in the objective
space.

In addition, a mutation operator is used in \cite{coello2004} to enhance
exploration of the search space in the beginning of the search. The mutation
rate is made zero as the algorithm converges.

With this background, we now describe modifications made in the
proposed MOPSO+ algorithm.
\begin{list}{(\alph{cntr2})}{\usecounter{cntr2}}
 \item
  Archive manipulation: The hypergrid approach used in
  \cite{coello2004} was modified in \cite{patil2018} to avoid changing
  of grid boundaries and for more efficient use of memory. This
  new approach, which uses a hypergrid with a fixed cell size and
  does not involve grid boundaries, is used in the MOPSO+ program.
 \item
  Leader selection: The leader selection process in MOPSO\,\cite{coello2004}
  is illustrated in
  Fig.~\ref{fig_gbest}\,(a).
  In each PSO iteration, each particle is assigned one of the ND
  particles in the archive, preferring less crowded hypercubes.
  In MOPSO+, we continue to use the roulette-wheel selection procedure
  of the MOPSO algorithm. However, to intensify exploration of the
  less-crowded regions of the archive, we assign the same leader to
  all particles, as shown in
  Fig.~\ref{fig_gbest}\,(b) and keep the same leader for
  $N_{\rm{leader}}^{\rm{const}}$ iterations.
 \item
  Mutation: In PSO, when the velocity and position update steps fail
  to generate new ND solutions, mutation can be useful\,\cite{parsopoulos2008}.
  In the context of the WDS benchmark problems, we have observed that
  there is an initial phase of MOPSO in which the ND set is improved
  relatively rapidly. However, beyond a certain point, the rate of
  generation of new solutions drops significantly. For this reason,
  different mutation schemes have been implemented in MOPSO+
  (see Fig.~\ref{fig_mut}).

  In the ``constant" option, the mutation probability remains constant
  (a low value such as 0.01). In the ``pulse" option, the probability
  is made non-zero only for $N_{\rm{mut}}$ iterations in the early
  stages and zero otherwise. In the ``periodic" option, the probability
  is made non-zero for $N_{\rm{mut}}$ iterations in every
  $N_{\rm{period}}$ iterations, thus periodically encouraging
  enhanced exploration. The mutation process itself is common in
  the three cases and involves changing one of the decision
  variables of the particle randomly.
 \item
  Local search: The local search operation has been described in
  Sec.~\ref{sec_ls}. We will refer to that procedure as a ``unit
  local search" (ULS) step. In MOPSO+, local search is implemented
  as follows.
  \begin{list}{(\roman{cntr2})}{\usecounter{cntr2}}
   \item
    As seen in Sec.~\ref{sec_ls}, a ULS step can lead to some improvement
    in the ND set. If it is applied again on the new ND set, further
    improvement is possible\,\cite{maler2016}. For this purpose, MOPSO+ allows 
    the ULS to be repeated $N_{LS}^{\rm{max}}$ times at a given PSO
    iteration. If, after some ULS steps, it is found that no further
    generation of new ND solutions is taking place, the LS step is
    discontinued.
   \item
    As explained in Sec.~\ref{sec_ls}, local search is expensive, and it
    is not practical to perform it in every PSO iteration. In MOPSO+,
    therefore, LS is performed periodically instead of every iteration.
    Furthermore, it was observed in the context of the benchmark WDS
    problems that LS is more effective in the early stages. Based on this
    observation, a two-stage LS strategy is implemented. From iteration
    $N_{PSO}^{(1)}$ to
    $N_{PSO}^{(2)}$, LS is performed every
    $T_1$ iterations, and after
    $N_{PSO}^{(2)}$, it is performed every
    $T_2$ iterations.
  \end{list}
\end{list}

\section{Results for medium problems}
\label{sec_medium}
In this section, we present results obtained with the MOPSO+ algorithm
for the ``medium" problems (MP) listed in Table~\ref{tbl_wang1}. We
first consider the HAN network to understand the effect of the various
algorithm parameters. We then choose the set of parameters which gives
the best performance for the HAN network and use it for the other three
benchmark problems in the MP
category, viz., the NYT, BLA, and GOY networks. We will refer to the ND
sets reported in \cite{coello2004} as ``UExeter PFs" and the ND sets given
by MOPSO+ as the ``MOPSO+ PFs."

\begin{table*}[!ht]
   \centering
   \caption{Comparison of UExeter and MOPSO+ PFs for the HAN network for
    different MOPSO+ options.}
    \hspace*{0cm}
    \begin{tabular}{|l|r|r|r|r|r|r|r|r|r|r|r|r|}
    \hline
    \multicolumn{1}{|c|}{\multirow{2}{*}{MOPSO+}}
    & {\multirow{2}{*}{$N_{PSO}^{\mathrm{max}}$}}
    & \multicolumn{5}{c|}{\B \T UExeter (PF-A)}
    & \multicolumn{5}{c|}{MOPSO+ (PF-B)}
    & {\multirow{2}{*}{$N^c$}}
    \\ \cline{3-12}
    \multicolumn{1}{|c|}{options}
    &  {}
    &  \multicolumn{1}{c|}{\B\T $N_A^t$}
    &  \multicolumn{1}{c|}{\B\T $N_A^a$}
    &  \multicolumn{1}{c|}{\B\T $N_A^u$}
    &  \multicolumn{1}{c|}{\B\T $N_A^r$}
    &  \multicolumn{1}{c|}{\B\T $N_{FE}^{\rm{net}}$}
    &  \multicolumn{1}{c|}{\B\T $N_B^t$}
    &  \multicolumn{1}{c|}{\B\T $N_B^a$}
    &  \multicolumn{1}{c|}{\B\T $N_B^u$}
    &  \multicolumn{1}{c|}{\B\T $N_B^r$}
    &  \multicolumn{1}{c|}{\B\T $N_{FE}^{\rm{net}}$}
    & {}
    \\ \hline
    \BB \TT LS/Leader$^{\rm{new}}$/M$^{\rm{periodic}}$
    & 10,000 & 575 & 534 &  1  & 41  & 90\,M & 750 & 748 & 215 &  2  & 74.6\,M & 533
    \\ \hline
    \BB \TT LS/Leader$^{\rm{old}}$/M$^{\rm{periodic}}$
    & 10,000 & 575 & 537 &  4  & 38  & 90\,M & 726 & 714 & 181 & 12  & 71.3\,M & 533
    \\ \hline
    \BB \TT LS/Leader$^{\rm{new}}$/M$^{\rm{none}}$
    & 10,000 & 575 & 552 &  88 & 23  & 90\,M & 629 & 581 & 117 & 48  & 59.6\,M & 464
    \\ \hline
    \BB \TT LS/Leader$^{\rm{new}}$/M$^{\rm{constant}}$
    & 10,000 & 575 & 536 &  13 & 39  & 90\,M & 721 & 703 & 180 & 18  & 69.1\,M & 523
    \\ \hline
    \BB \TT LS/Leader$^{\rm{new}}$/M$^{\rm{pulse}}$
    & 10,000 & 575 & 535 &  13 & 40  & 90\,M & 748 & 729 & 207 & 19  & 71.0\,M & 522
    \\ \hline
    \BB \TT LS$^{\rm{none}}$/Leader$^{\rm{new}}$/M$^{\rm{periodic}}$
    & 10,000 & 575 & 558 & 187 & 17  & 90\,M & 561 & 452 &  81 & 109 & 40.0\,M & 371
    \\ \hline
    \BB \TT LS$^{\rm{none}}$/Leader$^{\rm{new}}$/M$^{\rm{periodic}}$
    & 20,000 & 575 & 553 & 92 & 22  & 90\,M & 660 & 577 &  116 & 83 & 80.0\,M & 461
    \\ \hline
    \end{tabular}
\label{tbl_han}
\end{table*}

Table~\ref{tbl_han} shows the results (i.e.,
$N_A^t$,
$N_A^u$, etc. as defined in Sec.~\ref{sec_compare_pf}) for the HAN network with
$N_p \,$=$\, 200$ (number of particles).
For the first six rows of the table, the number of PSO iterations
$N_r \,$=$\, 10,000$ while for the last row, it is 20,000.

\begin{figure}[!ht]
\centering
\scalebox{0.9}{\includegraphics{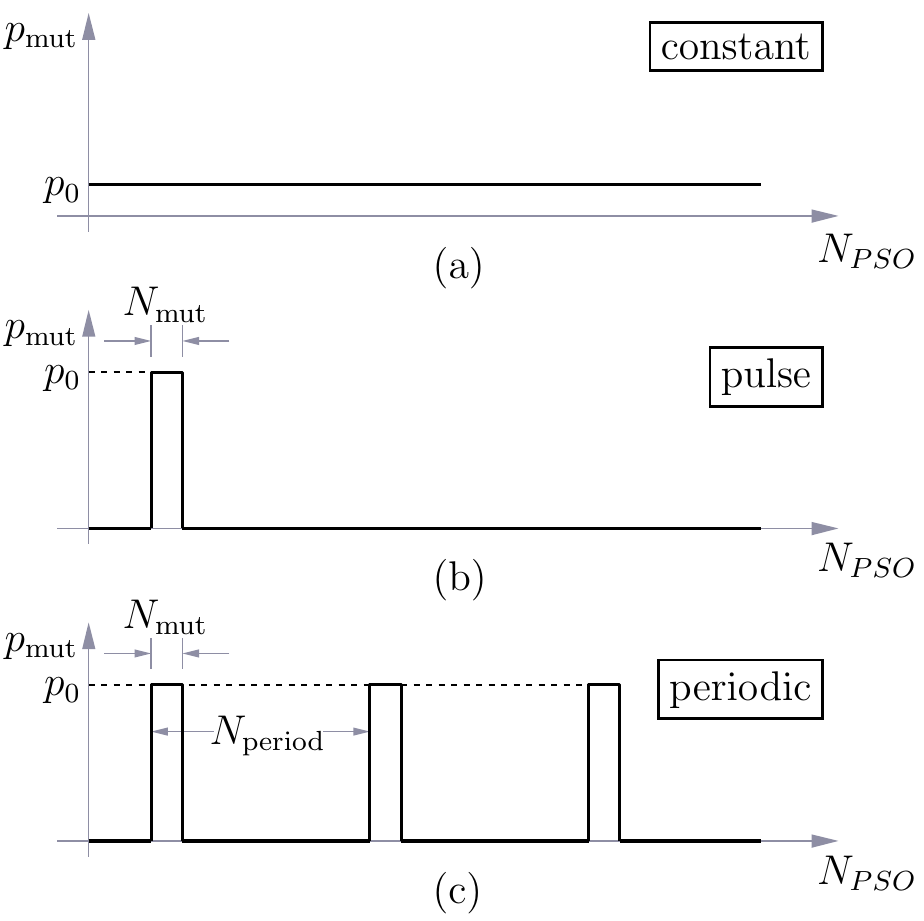}}
\caption{Mutation probability versus PSO iteration number for
different mutation schemes implemented in MOPSO+.}
\label{fig_mut}
\end{figure}

The MOPSO+ options mentioned in the first column have the
following meaning.
\begin{list}{(\alph{cntr2})}{\usecounter{cntr2}}
 \item
  LS: Local search is performed
  with a period of 100 between $N_{PSO} \,$=$\, 1,000$ and 5,000,
  and with a period of 1,000 thereafter.
  The parameter $N_{LS}^{\rm{max}}$ (see Sec.~\ref{sec_ls}) is set
  to 50.
 \item
  Leader$^{\rm{new}}$: The new leader assignment scheme
  (Fig.~\ref{fig_gbest}\,(b))
  is used with $N_{\rm{leader}}^{\rm{const}} \,$=$\, 10$.
 \item
  Leader$^{\rm{old}}$: The leader assignment scheme of
  \cite{coello2004}
  (Fig.~\ref{fig_gbest}\,(a))
  is used.
 \item
  M$^{\rm{none}}$: No mutation is performed.
 \item
  M$^{\rm{constant}}$: Constant mutation
  (Fig.~\ref{fig_mut}\,(a)) is performed with $p_0 \,$=$\, 0.02$.
 \item
  M$^{\rm{pulse}}$: Pulse mutation
  (Fig.~\ref{fig_mut}\,(b)) is performed with $p_0 \,$=$\, 1$,
  $N_{\mathrm{mut}} \,$=$\, 20$,
  starting with
  $N_{PSO} \,$=$\, 1,000$.
 \item
  M$^{\rm{periodic}}$: Periodic mutation
  (Fig.~\ref{fig_mut}\,(c)) is performed with $p_0 \,$=$\, 1$,
  $N_{\mathrm{mut}} \,$=$\, 20$,
  $N_{\mathrm{mut}}^{\mathrm{period}} \,$=$\, 1,000$,
  starting with
  $N_{PSO} \,$=$\, 1,000$.
 \item
  LS$^{\rm{none}}$: Local search is not used.
\end{list}

In each case, 20 independent runs of MOPSO+ were performed, and the
total number of function evaluations (using the EPANET program)
$N_{FE}^{\rm{net}}$ over all independent runs was recorded. We can
make the following observations from Table~\ref{tbl_han}.
\begin{list}{(\alph{cntr2})}{\usecounter{cntr2}}
 \item
  For each set of MOPSO+ options, a substantial number of new
  ND solutions (given by $N_B^u$) are found.
 \item
  Comparing the first two rows, we find that the new leader assignment
  scheme used in MOPSO+ gives a larger
  $N_B^u$. It also results in a smaller
  $N_A^u$, which is desirable, as explained in Sec.~\ref{sec_ls}.
 \item
  Among the different mutation schemes (see rows 1, 3, 4, 5), the periodic
  scheme gives the best PF.
 \item
  Local search plays a very important role (see rows 1 and 6) in MOPSO+.
  Without LS, we see that a large number of UExeter solutions
  ($N_A^u \,$=$\, 187$) are missed out by MOPSO+. Also, there is a
  substantial reduction (from 215 to 81) in the number of new solutions
  found by MOPSO+ when LS is not used.
 \item
  Although LS has improved the ND set, it has taken a larger number of
  function evaluations
  ($N_{FE}^{\rm{net}} \,$=$\, 74.6\,$M with LS
  as against
  $N_{FE}^{\rm{net}} \,$=$\, 40\,$M without LS, see rows 1 and 6).
  To make a fair comparison, we have run MOPSO+ without LS with a larger
  $N_{FE}^{\rm{net}} \,$=$\, 80\,$M by doubling
  $N_{PSO}^{\mathrm{max}}$, the number of PSO iterations.
  In this situation (row 7),
  $N_A^u$ and
  $N_B^u$ are worse than row 1 which implies that improvement due to LS
  is not simply because of increased
  $N_{FE}^{\rm{net}}$.
\end{list}

\begin{figure}[!ht]
\centering
\scalebox{0.8}{\includegraphics{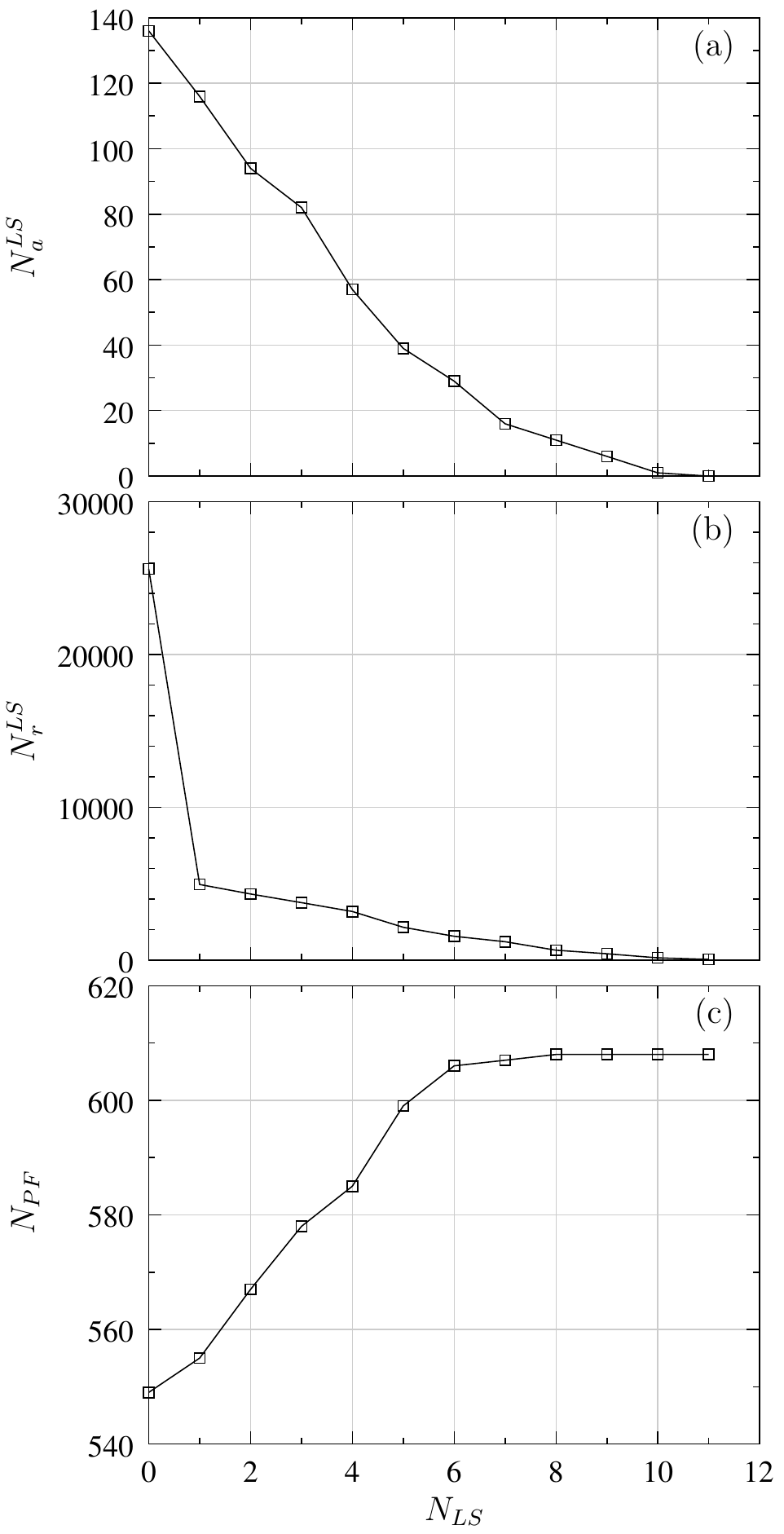}}
\caption{Illustration of the local search process at a specific
 PSO iteration ($N_{PSO} \,$=$\, 1,100$) for the HAN network with
 algorithm parameters same as those for row 1 of Table~\ref{tbl_han}.
 (a)\,Number of accepted solutions,
 (b)\,Number of rejected solutions,
 (c)\,Number of solutions in the ND set.}
\label{fig_han_ls1}
\end{figure}

It is instructive to look at the LS process for one specific run for the
first row of Table~\ref{tbl_han}. At a given PSO iteration at an early stage
($N_{PSO} \,$=$\, 1,100$),
the number of accepted solutions ($N_a^{LS}$),
the number of rejected solutions ($N_r^{LS}$), and the number of solutions
in the updated ND set ($N_{PF}$) are plotted versus
$N_{LS}$, the LS iteration number, in 
Figs.~\ref{fig_han_ls1}\,(a)-(c).
At the first LS iteration
($N_{LS} \,$=$\, 0$), the number of solutions in the ND set is about 550,
which means that there are $550\times 2\times 34$ (where 34 is the number
of decision variables), i.e., 37,400 neighbours if all of the 550 points are
interior. Some of these points would lie on the boundaries of the decision
space, and they would have less than $2\times 34$ neighbours each. Nevertheless,
it is clear that the number of solutions to be evaluated at the first LS iteration
is rather large. Of these, only 138 solutions
(Fig.~\ref{fig_han_ls1}\,(a))
get accepted in the ND set. About 25,000
(Fig.~\ref{fig_han_ls1}\,(b))
are found to be dominated by other solutions in the ND set and are therefore
rejected.

During the first LS iteration (i.e., the ULS step described earlier),
a list of all solutions being evaluated is
prepared using the tree format shown in Fig.~\ref{fig_tree}. In the
subsequent LS iterations, the neighbours of the updated ND solutions are
compared against this list, and only those not found in the list are
evaluated. Therefore, a much smaller number of solutions need to be tried
in the subsequent LS iterations. For example, for
$N_{LS} \,$=$\, 1$,
only about 5,000 solutions need to be evaluated. This process continues until
either
$N_{LS}^{\rm{max}}$ iterations are completed or when an LS iteration
fails to generate any new ND solutions. For the example shown in
Fig.~\ref{fig_han_ls1}, the latter situation is reached (at
$N_{LS} \,$=$\, 11$). With each LS iteration, the ND set gets improved, and in
this specific example, the number of solutions in the ND set also increases
(Fig.~\ref{fig_han_ls1}\,(c)) with each LS iteration.

\begin{figure}[!ht]
\centering
\scalebox{0.8}{\includegraphics{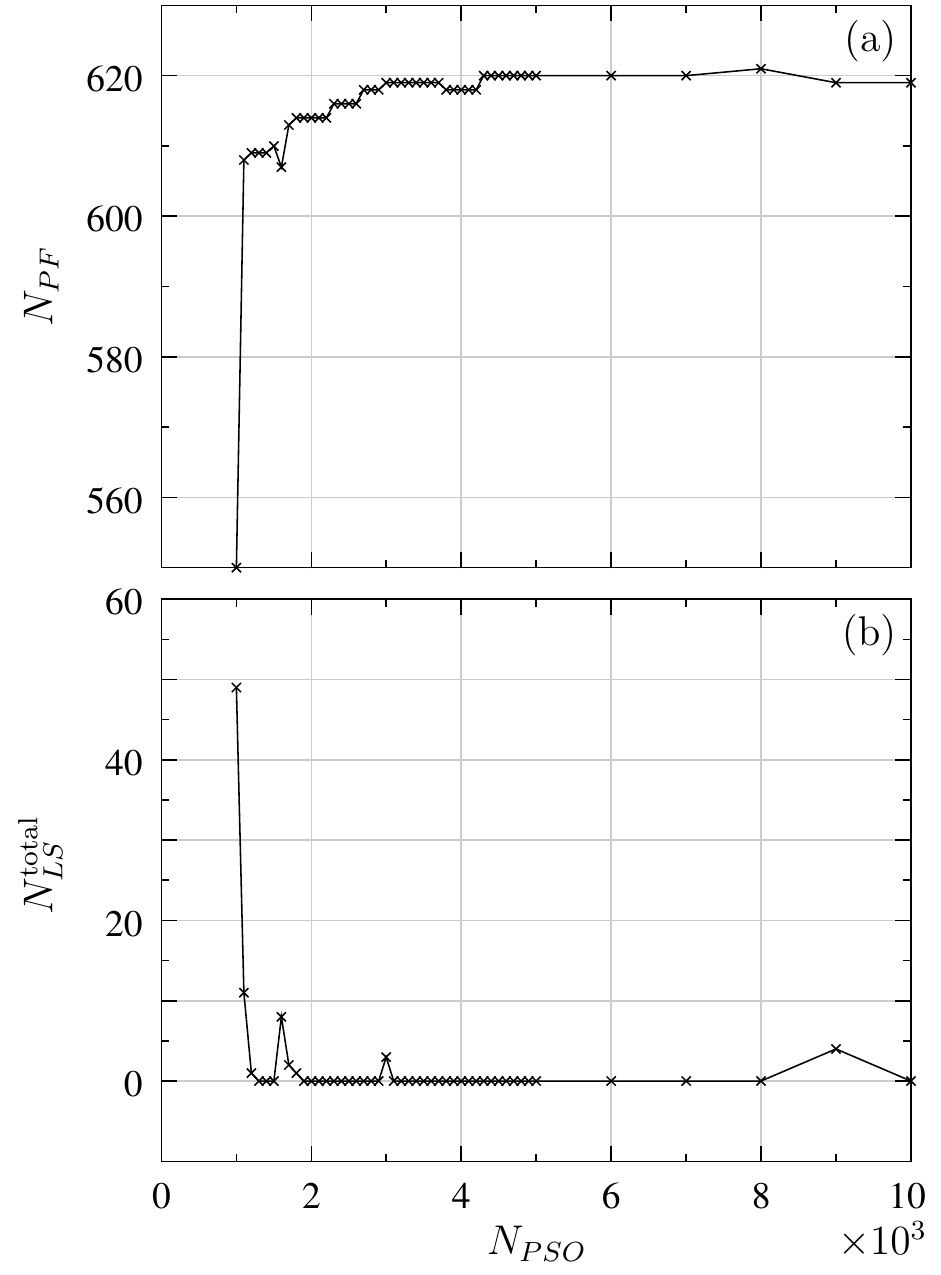}}
\caption{(a)\,Number of ND solutions ($N_{PF}$) and (b)\,number of
 total LS iterations ($N_{LS}^{\mathrm{total}}$) versus PSO iteration
 number for the HAN network with
 algorithm parameters same as those for row 1 of Table~\ref{tbl_han}.}
\label{fig_han_ls2}
\end{figure}

Fig.~\ref{fig_han_ls2}\,(a) shows the variation of the size of the ND set
($N_{PF}$), and 
Fig.~\ref{fig_han_ls2}\,(b) shows the total number of LS iterations tried
($N_{LS}^{\mathrm{total}}$)
as the MOPSO+ algorithm proceeds. Note that, as mentioned earlier, the period
of local search is low (100) up to
$N_{PSO} \,$=$\, 5,000$ and high
(1,000) thereafter, and that is the reason for the larger
density of data points up to
$N_{PSO} \,$=$\, 5,000$ in
Fig.~\ref{fig_han_ls2}\,(b).
For the first LS step
(at $N_{PSO} \,$=$\, 1,000$),
$N_{LS}^{\mathrm{total}} \,$=$\, N_{LS}^{\mathrm{max}} \,$=$\, 50$.
Beyond this point, generation of ND solutions due to LS slows down, and after
about 2,000 PSO iterations, we see that
$N_{LS}^{\mathrm{total}}$ mostly remains zero (which means only one LS iteration
was tried which failed to produce new ND points). This points to a side benefit
of using LS: it provides a necessary (but not sufficient) condition for convergence
of an MOEA. If LS yields additional ND points, we can say that the MOEA has not
converged.

\begin{table*}[!ht]
   \centering
   \caption{Comparison of UExeter and MOPSO+ PFs for different values of
    $N_p$ (number of particles), $N_{PSO}^{\mathrm{max}}$ (number of PSO
    iterations), and $N_r$ (number of independent runs) for the NYT network.}
    \hspace*{0cm}
    \begin{tabular}{|c|r|c|r|r|r|r|r|r|r|r|r|r|r|}
    \hline
    \multicolumn{3}{|c|}{\B \T MOPSO+ options}
    & \multicolumn{5}{c|}{\B \T UExeter (PF-A)}
    & \multicolumn{5}{c|}{MOPSO+ (PF-B)}
    & {\multirow{2}{*}{$N^c$}}
    \\ \cline{1-13}
       \multicolumn{1}{|c|}{\B\T $N_p$}
    &  \multicolumn{1}{c|}{\B\T $N_{PSO}^{\mathrm{max}}$}
    &  \multicolumn{1}{c|}{\B\T $N_r$}
    &  \multicolumn{1}{c|}{\B\T $N_A^t$}
    &  \multicolumn{1}{c|}{\B\T $N_A^a$}
    &  \multicolumn{1}{c|}{\B\T $N_A^u$}
    &  \multicolumn{1}{c|}{\B\T $N_A^r$}
    &  \multicolumn{1}{c|}{\B\T $N_{FE}^{\mathrm{net}}$}
    &  \multicolumn{1}{c|}{\B\T $N_B^t$}
    &  \multicolumn{1}{c|}{\B\T $N_B^a$}
    &  \multicolumn{1}{c|}{\B\T $N_B^u$}
    &  \multicolumn{1}{c|}{\B\T $N_B^r$}
    &  \multicolumn{1}{c|}{\B\T $N_{FE}^{\mathrm{net}}$}
    & {}
    \\ \hline
    \B \T  200 & 20,000 & 30 & 627 & 599 & 20  & 28  & 90\,M & 648 & 639 & 60  &  9  & 150.7\,M & 579
    \\ \hline
    \B \T  300 & 20,000 & 30 & 627 & 597 & 28  & 30  & 90\,M & 670 & 630 & 61  & 40  & 211.3\,M & 569
    \\ \hline
    \B \T  100 & 60,000 & 10 & 627 & 596 & 25  & 31  & 90\,M & 655 & 635 & 64  & 20  &  76.7\,M & 571
    \\ \hline
    \B \T  100 & 60,000 & 17 & 627 & 595 &  4  & 32  & 90\,M & 661 & 656 & 65  &  5  & 130.3\,M & 591
    \\ \hline
    \end{tabular}
\label{tbl_nyt}
\end{table*}

\begin{table*}[!ht]
   \centering
   \caption{Comparison of UExeter and MOPSO+ PFs for the four MP networks
    described in \cite{coello2004}. The MOPSO+ algorithm parameters are the
    same as row 1 of Table~\ref{tbl_han}.}
    \hspace*{0cm}
    \begin{tabular}{|c|c|r|c|r|r|r|r|r|r|r|r|r|r|r|}
    \hline
      {\multirow{2}{*}{Network}}
    & \multicolumn{3}{|c|}{\B \T MOPSO+ options}
    & \multicolumn{5}{c|}{\B \T UExeter (PF-A)}
    & \multicolumn{5}{c|}{MOPSO+ (PF-B)}
    & {\multirow{2}{*}{$N^c$}}
    \\ \cline{2-14}
       {}
    &  \multicolumn{1}{|c|}{\B\T $N_p$}
    &  \multicolumn{1}{c|}{\B\T $N_{PSO}^{\mathrm{max}}$}
    &  \multicolumn{1}{c|}{\B\T $N_r$}
    &  \multicolumn{1}{c|}{\B\T $N_A^t$}
    &  \multicolumn{1}{c|}{\B\T $N_A^a$}
    &  \multicolumn{1}{c|}{\B\T $N_A^u$}
    &  \multicolumn{1}{c|}{\B\T $N_A^r$}
    &  \multicolumn{1}{c|}{\B\T $N_{FE}^{\mathrm{net}}$}
    &  \multicolumn{1}{c|}{\B\T $N_B^t$}
    &  \multicolumn{1}{c|}{\B\T $N_B^a$}
    &  \multicolumn{1}{c|}{\B\T $N_B^u$}
    &  \multicolumn{1}{c|}{\B\T $N_B^r$}
    &  \multicolumn{1}{c|}{\B\T $N_{FE}^{\mathrm{net}}$}
    & {}
    \\ \hline
    \B \T NYT & 100 & 60,000 & 17 & 627 & 595 & 4~ & 32 & 90\,M &  661 &  656 &  65 & 5~ & 130.3\,M & 591
    \\ \hline
    \B \T BLA & 200 & 10,000 & 10 & 901 & 849 & 0~ & 52 & 90\,M & 1045 & 1045 & 196 & 0~ &  44.1\,M & 849
    \\ \hline
    \B \T HAN & 200 & 10,000 & 20 & 575 & 534 & 1~ & 41 & 90\,M &  750 &  748 & 215 & 2~ &  74.6\,M & 533
    \\ \hline
    \B \T GOY & 100 & 20,000 & 10 & 489 & 444 & 3~ & 45 & 90\,M &  571 &  570 & 129 & 1~ &  37.9\,M & 441
    \\ \hline
    \end{tabular}
\label{tbl_mp_all}
\end{table*}

We now use the MOPSO+ algorithm parameters which have been found to work
well for the HAN problem (option
LS/Leader$^{\rm{new}}$/M$^{\rm{periodic}}$
in Table~\ref{tbl_han}) for optimisation of the other medium problems described in
\cite{coello2004}, viz., the NYT, BLA, and GOY networks. The results for the
NYT network are shown in Table~\ref{tbl_nyt}. We observe that the $N_B^u$ values
are comparable in the four cases listed in the table. However, for
$N_p \,$=$\, 100$,
$N_{PSO}^{\mathrm{max}} \,$=$\, 60,000$, and
$N_r \,$=$\, 17$,
$N_A^u$ is substantially lower than the other three cases.

Comparing rows 1 and 2 of the table, we conclude that increasing the number of
particles is not necessarily beneficial, which was also pointed out in
\cite{wang2014} for other MOEAs. A systematic study of the effect of
$N_p$ and $N_{PSO}^{\mathrm{max}}$ on the performance of MOPSO+ has not been
carried out in this work. Instead, we present the results obtained after limited
experimentation with these parameters.

Table~\ref{tbl_mp_all} shows the results for the four MP benchmark problems
described in \cite{coello2004}. In each case, a substantial number of new ND solutions
($N_B^u$) have been found by MOPSO+. Furthermore, almost all ND solutions in the UExeter
PFs are covered by MOPSO+ (see the $N_A^u$ column in the table). The improvement in the PFs
have been obtained with a smaller
$N_{FE}^{\mathrm{net}}$
for three networks and with a larger
$N_{FE}^{\mathrm{net}}$
for the NYT network. These results overall point to the usefulness of the MOPSO+
approach for the WDS design problem.

\begin{figure*}
\centering
\scalebox{0.9}{\includegraphics{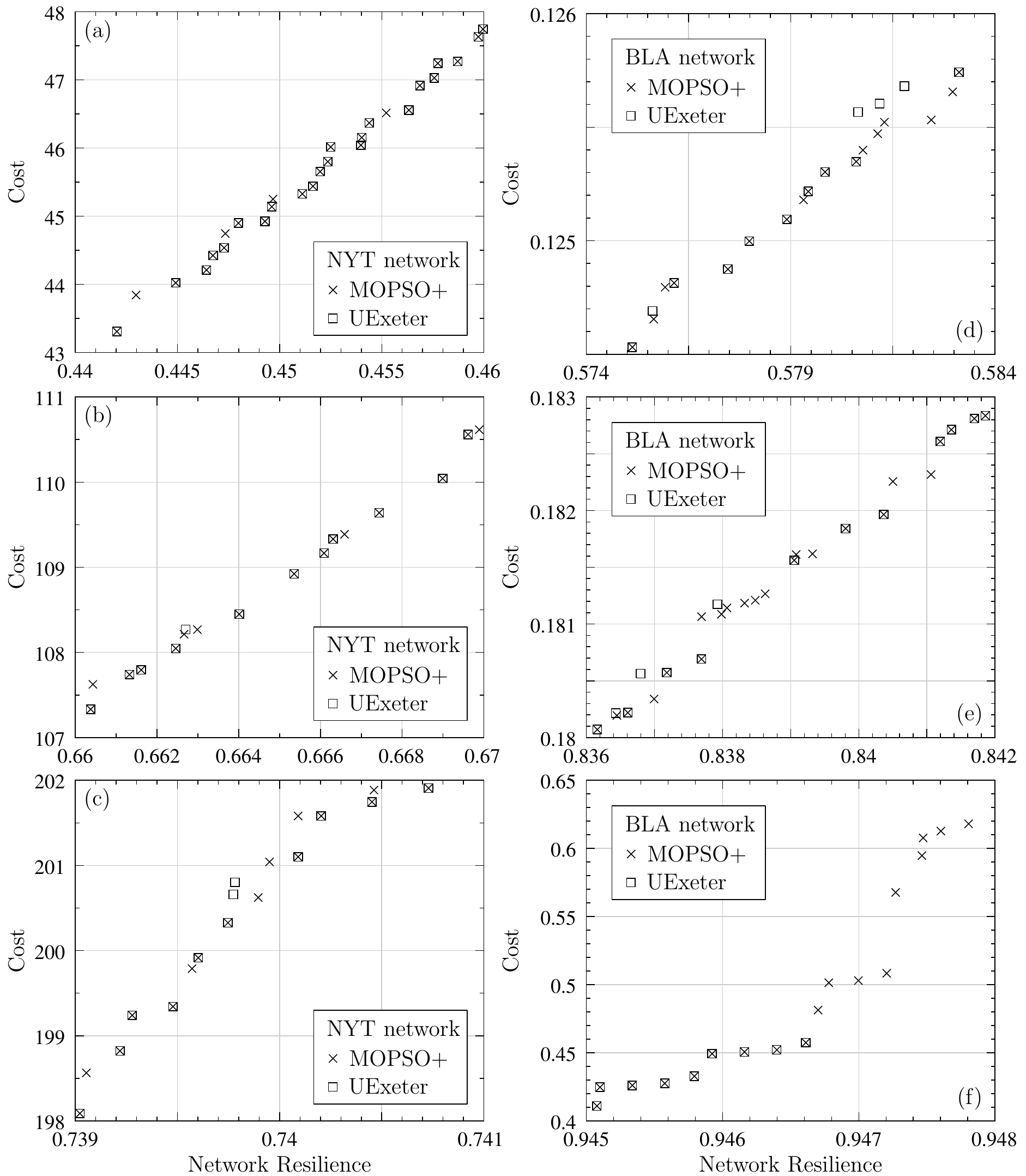}}
\caption{UExeter\,\cite{exeter} and MOPSO+ ND sets for the NYT and BLA networks for
different ranges of network resilience.}
\label{fig_nyt_bla}
\end{figure*}

\begin{figure*}
\centering
\scalebox{0.9}{\includegraphics{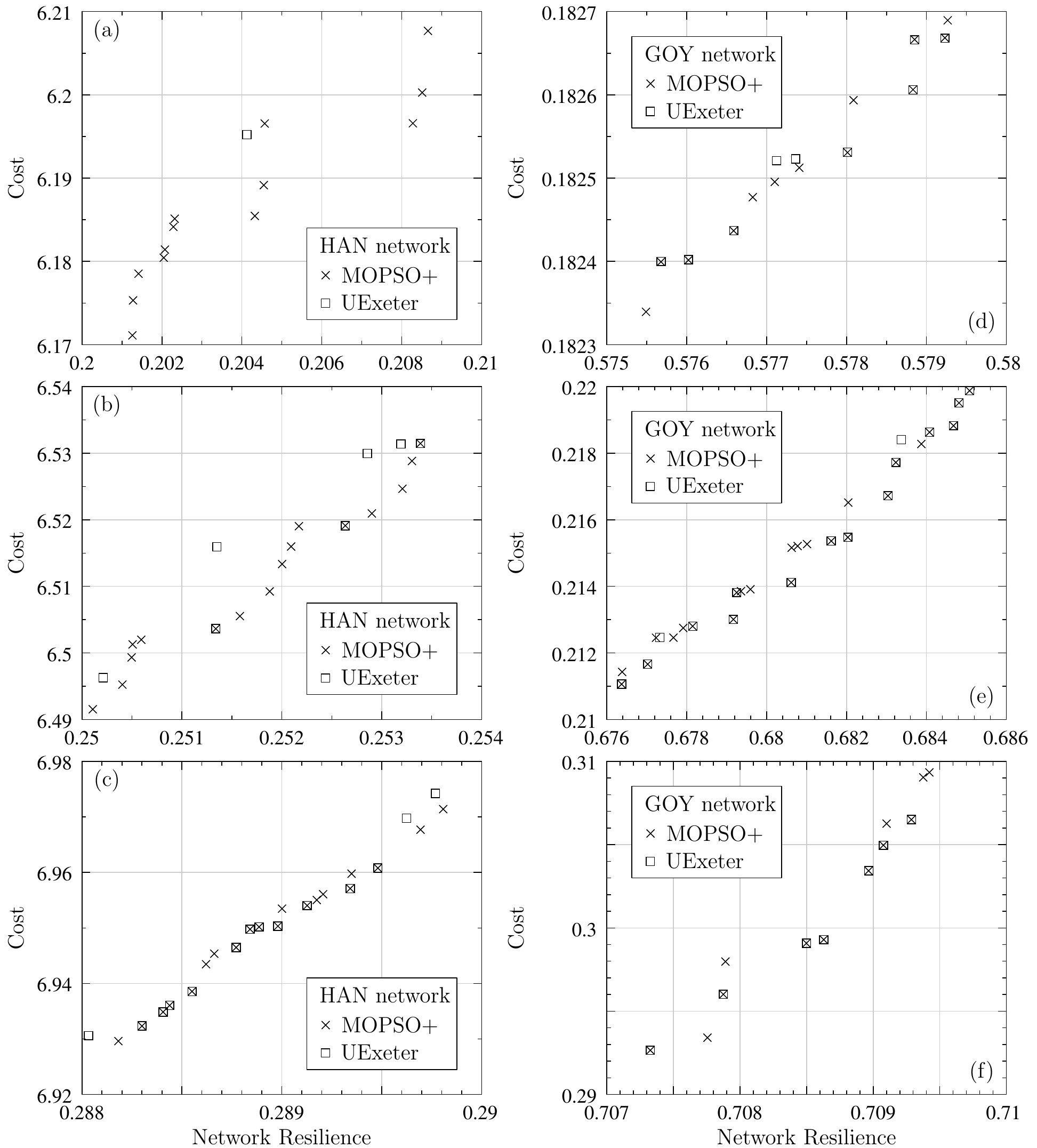}}
\caption{UExeter\,\cite{exeter} and MOPSO+ ND sets for the HAN and GOY networks for
different ranges of network resilience.}
\label{fig_han_goy}
\end{figure*}

Figs.~\ref{fig_nyt_bla}\,(a)-(f) and
\ref{fig_han_goy}\,(a)-(f) show expanded plots of the UExeter and MOPSO+
PFs. It can be seen that some of the MOPSO+ solutions dominate some of the UExeter
solutions.

\begin{table*}[!ht]
\centering
\caption{Objective function and decision variable values for a few solutions
of Fig.~\ref{fig_nyt_bla}\,(d) obtained by MOPSO+ for the BLA network.}
\hspace*{0cm}
\begin{tabular}{|c|r|r|r|r|r|r|r|r|r|r|r|r|r|r|}
\hline
{}
& \multicolumn{1}{c|}{\B\T sol.\,1}
& \multicolumn{1}{c|}{\B\T sol.\,2}
& \multicolumn{1}{c|}{\B\T sol.\,3}
& \multicolumn{1}{c|}{\B\T sol.\,4}
& \multicolumn{1}{c|}{\B\T sol.\,5}
& \multicolumn{1}{c|}{\B\T sol.\,6}
& \multicolumn{1}{c|}{\B\T sol.\,7}
\\ \hline
\B\T Resilience & 0.57931 & 0.57943 & 0.57984 & 0.58060 & 0.58077 & 0.58113 & 0.58129
\\ \hline
\B\T Cost       & 0.12518 & 0.12522 & 0.12530 & 0.12535 & 0.12540 & 0.12547 & 0.12552
\\ \hline
\B\T $d_1$      & 254.0 & 254.0 & 254.0 & 254.0 & 254.0 & 254.0 & 254.0
\\ \hline
\B\T $d_2$      & 101.6 & 101.6 & 101.6 & 101.6 & 101.6 & 101.6 & 101.6
\\ \hline
\B\T $d_3$      & 203.2 & 203.2 & 203.2 & 203.2 & 203.2 & 203.2 & 203.2
\\ \hline
\B\T $d_4$      & 203.2 & 203.2 & 203.2 & 203.2 & 203.2 & 203.2 & 203.2
\\ \hline
\B\T $d_5$      & 152.4 & 152.4 & 152.4 & 152.4 & 152.4 & 152.4 & 152.4
\\ \hline
\B\T $d_6$      & 152.4 & 152.4 & 152.4 & 152.4 & 152.4 & 152.4 & 152.4
\\ \hline
\B\T $d_7$      & 152.4 & 152.4 & 152.4 & 152.4 & 152.4 & 152.4 & 152.4
\\ \hline
\B\T $d_8$      &  25.4 &  25.4 &  25.4 &  25.4 &  25.4 &  25.4 &  25.4
\\ \hline
\B\T $d_9$      & 152.4 & 152.4 & 152.4 & 152.4 & 152.4 & 152.4 & 152.4
\\ \hline
\B\T $d_{10}$   & 101.6 & 101.6 & 101.6 & 101.6 & 101.6 & 101.6 & 101.6
\\ \hline
\B\T $d_{11}$   & 101.6 &  76.2 & 101.6 &  76.2 &  76.2 &  76.2 &  76.2
\\ \hline
\B\T $d_{12}$   & 152.4 & 152.4 & 152.4 & 152.4 & 152.4 & 152.4 & 152.4
\\ \hline
\B\T $d_{13}$   & 152.4 & 152.4 & 152.4 & 152.4 & 152.4 & 152.4 & 152.4
\\ \hline
\B\T $d_{14}$   & 152.4 & 152.4 & 152.4 & 152.4 & 152.4 & 152.4 & 152.4
\\ \hline
\B\T $d_{15}$   &  76.2 & 101.6 & 101.6 &  76.2 &  76.2 & 101.6 & 101.6
\\ \hline
\B\T $d_{16}$   &  50.8 &  50.8 &  50.8 &  50.8 &  50.8 &  50.8 &  50.8
\\ \hline
\B\T $d_{17}$   &  50.8 &  50.8 &  50.8 &  50.8 &  50.8 &  50.8 &  50.8
\\ \hline
\B\T $d_{18}$   &  76.2 &  50.8 &  76.2 &  25.4 &  76.2 &  25.4 &  76.2
\\ \hline
\B\T $d_{19}$   &  25.4 &  25.4 &  25.4 &  25.4 &  25.4 &  25.4 &  25.4
\\ \hline
\B\T $d_{20}$   & 101.6 & 101.6 & 101.6 & 152.4 & 101.6 & 152.4 & 101.6
\\ \hline
\B\T $d_{21}$   &  25.4 &  25.4 &  25.4 &  25.4 &  25.4 &  25.4 &  25.4
\\ \hline
\B\T $d_{22}$   & 152.4 & 152.4 & 152.4 & 152.4 & 152.4 & 152.4 & 152.4
\\ \hline
\B\T $d_{23}$   & 101.6 & 152.4 & 101.6 & 101.6 & 152.4 & 101.6 & 152.4
\\ \hline
\end{tabular}
\label{tbl_bla}
\end{table*}

Even though many of the unique MOPSO+ solutions (i.e., solutions present in the
MOPSO+ PFs but not in the UExeter PFs) are close in the objective space to the
previously known UExeter solutions, they can differ substantially in the decision
space. For example, some of the ND solutions obtained by MOPSO+ for the BLA network
are shown in Table~\ref{tbl_bla}. Of these, solutions 2, 3, 4 are also present in the
UExeter PF while solutions 1, 5, 6, 7 are new. Although the cost and resilience values
for solutions 1 and 2 differ only in the fourth decimal place, we observe that these
two solutions vary substantially in the decision space~-- four diameter values, viz.,
$d_{11}$, $d_{15}$, $d_{18}$, $d_{23}$, are different.
This could have important implications in practice because of non-technical issues
such as land acquisition, and a wider choice of ND solutions may well open up
previously unknown options for the DM. The metrics proposed in Sec.~\ref{sec_compare_pf}
therefore seem to be more attractive from a practical perspective than metrics such as
average cost, lowest cost, standard deviation, or spread.

\section{Results for an intermediate problem}
\label{sec_inter}
As seen in Sec.~\ref{sec_ls}, local search is an expensive step, and with a
large number of decision variables (in our context, the number of pipes)
involved in the networks of IP and LP categories in \cite{wang2014},
the repeated ULS procedure
described for the MP category (see Sec.~\ref{sec_mopso}) can become
prohibitively expensive. In this section, we will consider one of the IPs,
the PES network (see Table~\ref{tbl_wang1}). Since there are 99 decision
variables, local search around one solution involves $99\times 2 \,$=$\, 198$
function evaluations. As we shall see, the ND set for this network is much
larger than the MP networks, making even one unit local search step involving
evaluation of all neighbours of all solutions in the ND set consume
significant computation time.

In view of the above difficulty, the ND set is only partially explored during
local search (see \cite{liefooghe2012},\cite{jaszkiewicz2018}),
and only one ULS step is performed. In particular,
$N_{ND}^{\mathrm{max}}$ solutions from the current ND set are randomly selected, and all
neighbours of the selected solutions are evaluated. Apart from this change,
other algorithm details remain the same as before.

For the PES problem, 10 independent runs of MOPSO+ were performed with
$N_p \,$=$\, 200$ and
$N_{PSO} \,$=$\, 1,000,000$. Periodic mutation was employed, with
$p_0 \,$=$\, 1$,
$N_{\mathrm{mut}} \,$=$\, 50$,
$N_{\mathrm{mut}}^{\mathrm{period}} \,$=$\, 1,000$,
starting with
$N_{PSO} \,$=$\, 1,000$.
The new scheme for finding the leader (see Sec.~\ref{sec_mopso}) is used, with
$N_{\rm{leader}}^{\rm{const}} \,$=$\, 10$.
Local search was performed with a period of 1,000 between
$N_{PSO} \,$=$\, 1,000$ and 10,000,
and with a period of 10,000 thereafter.
$N_{ND}^{\mathrm{max}}$ was set to 200, which means only 200 solutions
from the current ND set are randomly picked for any LS step.

\begin{figure}[!ht]
\centering
\scalebox{0.9}{\includegraphics{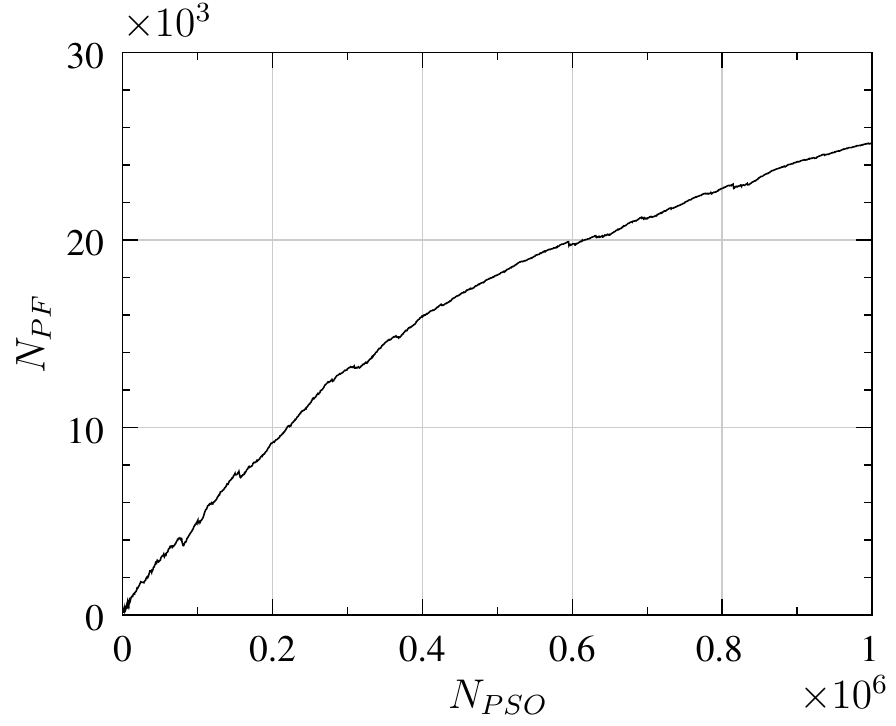}}
\caption{Number of ND solutions ($N_{PF}$) versus PSO iteration number
for the PES network.}
\label{fig_pes_nd_npso}
\end{figure}

Fig.~\ref{fig_pes_nd_npso} shows $N_{PF}$ (the number of ND solutions) versus
PSO iteration number for one independent run. A much larger $N_{PF}$
(about 25,000) is obtained for this network. Note that, in contrast to
Fig.~\ref{fig_han_ls2} for the HAN network, $N_{PF}$ has not saturated
in this case, indicating that $N_{PF}$ may increase further with additional
PSO iterations.

\begin{table*}[!ht]
\centering
   \caption{Comparison of UExeter and MOPSO+ PFs for different values of
    $N_r$ (number of independent runs) for the PES network.}
   \hspace*{0cm}
    \begin{tabular}{|c|r|r|r|r|r|r|r|r|r|r|r|}
    \hline
    {\multirow{2}{*}{\B\T $N_r$}}
    & \multicolumn{5}{c|}{\B \T UExeter (PF-A)}
    & \multicolumn{5}{c|}{MOPSO+ (PF-B)}
    & {\multirow{2}{*}{$N^c$}}
    \\ \cline{2-11}
    {}
    &  \multicolumn{1}{c|}{\B\T $N_A^t$}
    &  \multicolumn{1}{c|}{\B\T $N_A^a$}
    &  \multicolumn{1}{c|}{\B\T $N_A^u$}
    &  \multicolumn{1}{c|}{\B\T $N_A^r$}
    &  \multicolumn{1}{c|}{\B\T $N_{FE}^{\rm{net}}$}
    &  \multicolumn{1}{c|}{\B\T $N_B^t$}
    &  \multicolumn{1}{c|}{\B\T $N_B^a$}
    &  \multicolumn{1}{c|}{\B\T $N_B^u$}
    &  \multicolumn{1}{c|}{\B\T $N_B^r$}
    &  \multicolumn{1}{c|}{\B\T $N_{FE}^{\rm{net}}$}
    & {}
    \\ \hline
     \B\T 1 & 782 & 307 & 292 & 475 & 150\,M & 21,536 & 12,161 & 12,146 & 9,375 &   269\,M & 15~
    \\ \hline
     \B\T 2 & 782 & 190 & 175 & 592 & 150\,M & 20,736 & 14,856 & 14,841 & 5,880 &   538\,M & 15~
    \\ \hline
     \B\T 3 & 782 & 167 & 152 & 615 & 150\,M & 22,949 & 20,081 & 20,066 & 2,868 &   808\,M & 15~
    \\ \hline
     \B\T 4 & 782 & 166 & 152 & 616 & 150\,M & 22,962 & 20,094 & 20,080 & 2,868 & 1,077\,M & 14~
    \\ \hline
     \B\T 5 & 782 &  39 &  22 & 743 & 150\,M & 23,267 & 23,168 & 23,151 &    99 & 1,348\,M & 17~
    \\ \hline
     \B\T 6 & 782 &  37 &  19 & 745 & 150\,M & 24,643 & 24,547 & 24,529 &    96 & 1,619\,M & 18~
    \\ \hline
     \B\T 7 & 782 &  36 &  19 & 746 & 150\,M & 25,607 & 25,510 & 25,493 &    97 & 1,889\,M & 17~
    \\ \hline
     \B\T 8 & 782 &  36 &  19 & 746 & 150\,M & 26,145 & 26,048 & 26,031 &    97 & 2,160\,M & 17~
    \\ \hline
     \B\T 9 & 782 &  35 &  18 & 747 & 150\,M & 26,488 & 26,399 & 26,382 &    89 & 2,431\,M & 17~
    \\ \hline
    \B\T 10 & 782 &  35 &  18 & 747 & 150\,M & 30,267 & 30,178 & 30,161 &    89 & 2,703\,M & 17~
    \\ \hline
    \end{tabular}
\label{tbl_pes}
\end{table*}

Table~\ref{tbl_pes} shows the effect of $N_r$ (the number of independent runs)
on the performance of MOPSO+. With a single run,
$N_B^u \,$=$\, 12,146$ new solutions are generated. However,
$N_A^u \,$=$\, 292$ solutions from the UExeter PF are missed out by MOPSO+.
The number of function evaluations for this run is 269\,M out of which
$N_p\times N_r \,$=$\, 200$\,M are due to the PSO part and 69\,M are due
to local search.
This value of $N_{FE}^{\rm{net}}$ is already larger than the UExeter value
of 150\,M. As more independent runs are performed,
$N_{FE}^{\rm{net}}$ increases proportionately. However, the number of additional
ND solutions increases with $N_r$, and the number of UExeter ND solutions ($N_A^u$)
missed by MOPSO+ decreases, both factors being favourable. For ten runs,
$N_{FE}^{\rm{net}}$ for MOPSO+ is about 18 times larger than UExeter. Considering
the large number of new ND solutions (about 38 times larger than UExeter), the
additional computational effort seems worthwhile.

\begin{figure}[!ht]
\centering
\scalebox{0.9}{\includegraphics{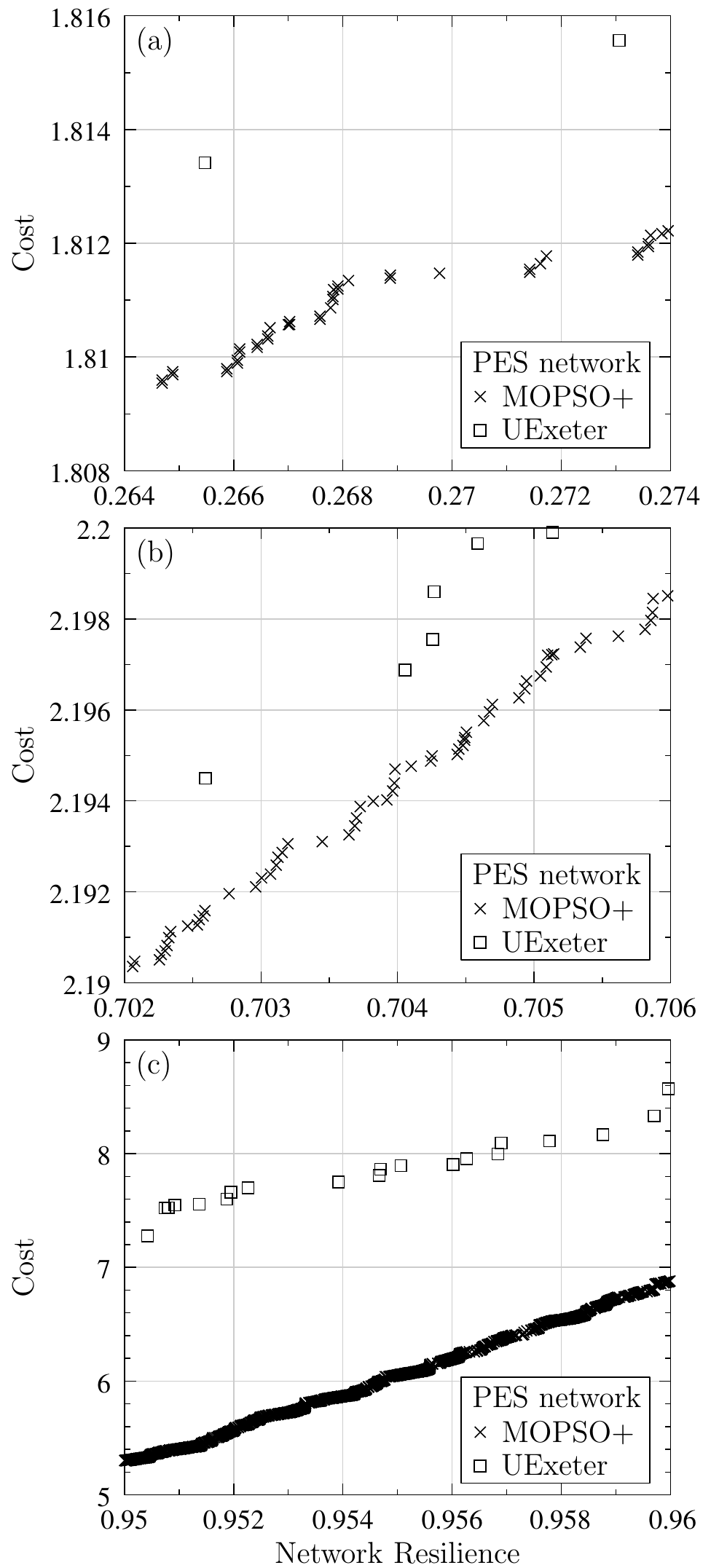}}
\caption{UExeter\,\cite{exeter} and MOPSO+ ND sets for the PES network for
different ranges of network resilience.}
\label{fig_pes}
\end{figure}

Some of the ND solutions obtained by MOPSO+ are shown in
Figs.~\ref{fig_pes}\,(a)-(c), along with UExeter solutions.
It can be seen that the density of solutions obtained by MOPSO+ is
larger. Also, for some values of network resilience (see
Fig.~\ref{fig_pes}\,(c)), the cost for some of the MOPSO+ solutions
is lower by as much as 25\,\%.

\section{Conclusions}
\label{sec_concl}
In summary, a hybrid algorithm (MOPSO+) comprising a multi-objective particle
swarm optimisation algorithm and local search is presented for water
distribution system design. Four medium and one
intermediate benchmark networks are considered, and for each of them,
a significant number of new non-dominated solutions have been obtained
using the proposed algorithm, thus improving the best-known Pareto
fronts for these problems. For medium networks, the total
number of function evaluations for MOPSO+ was of the same order
as in \cite{wang2014}.
For the intermediate network considered here, although the total $N_{FE}$ was
much larger (18 times) for MOPSO+, the large number of new solutions
generated by MOPSO+ justifies the additional computational effort.

Based on the results presented here, MOPSO+ seems to be an attractive
option for WDS design. Some important issues for future research in
this area are listed below.
\begin{list}{(\alph{cntr2})}{\usecounter{cntr2}}
 \item
  Performance of MOPSO+ for other networks in the IP and LP category
  needs to be investigated.
 \item
  A unified set of algorithmic parameters for all problems of a
  particular complexity (MP, IP, LP) is desirable. More work is
  required to explore this possibility.
 \item
  The LS mechanism introduced here can be applied to other multi-objective
  optimisation problems with discrete decision variables.
 \item
  In order to speed up convergence, techniques such as reduction of search
  space\,\cite{barlow2014}, high-quality initial population\,\cite{bi2016} could be
  combined with the MOPSO+ algorithm.
\end{list}

\bibliographystyle{IEEEtran}
\bibliography{arx1}

\vfill

\end{document}